\documentclass[runningheads]{llncs}
\usepackage{graphicx}

\usepackage{tikz}
\usepackage{comment} 
\usepackage{amsmath,amssymb} %
\usepackage{color}
\usepackage{bm}
\usepackage[font=small]{caption}

\begin{document}
\pagestyle{headings}
\mainmatter
\def\ECCVSubNumber{100}  %

\title{Motion Capture from Internet Videos} %

\titlerunning{Motion Capture from Internet Videos}
\author{Junting Dong\inst{1, \star} \and
Qing Shuai\inst{1,\star} \and
Yuanqing Zhang\inst{1} \and
Xian Liu\inst{1} \and \\
Xiaowei Zhou\inst{1} \and
Hujun Bao\inst{2,1}}
\authorrunning{J. Dong et al.}
\institute{$^1$ State Key Lab of CAD\&CG, Zhejiang University, $^2$ Zhejiang Lab
}
\maketitle

\begin{abstract}

Recent advances in image-based human pose estimation make it possible to capture 3D human motion from a single RGB video. However, the inherent depth ambiguity and self-occlusion in a single view prohibit the recovery of as high-quality motion as multi-view reconstruction. While multi-view videos are not common, the videos of a celebrity performing a specific action are usually abundant on the Internet. Even if these videos were recorded at different time instances, they would encode the same motion characteristics of the person. Therefore, we propose to capture human motion by jointly analyzing these Internet videos instead of using single videos separately. However, this new task poses many new challenges that cannot be addressed by existing methods, as the videos are unsynchronized, the camera viewpoints are unknown, the background scenes are different, and the human motions are not exactly the same among videos. To address these challenges, we propose a novel optimization-based framework and experimentally demonstrate its ability to recover much more precise and detailed motion from multiple videos, compared against monocular motion capture methods.{\renewcommand{\thefootnote}{\fnsymbol{footnote}} \footnotetext[1]{Authors contributed equally.}}


\end{abstract}

\section{Introduction}

Human motion capture (MoCap) is a core technology in a variety of applications such as movie production, video game development, sports analysis and interactive entertainment. While there have been some commercial solutions to MoCap, e.g., optical MoCap systems like Vicon, these systems are for professionals but not commodities. The systems are expensive and hard to calibrate. More importantly, the performers need to present in the studio to perform actions, which makes it impossible to collect large-scale motion data for a large population.
For example, producing an animated avatar of a celebrity needs to invite the person to the MoCap studio, which is not always feasible especially for amateur productions.

\begin{figure}[t]
	\centering
	\includegraphics[width=0.9\linewidth,trim={6cm 17.7cm 6.5cm 7cm},clip]{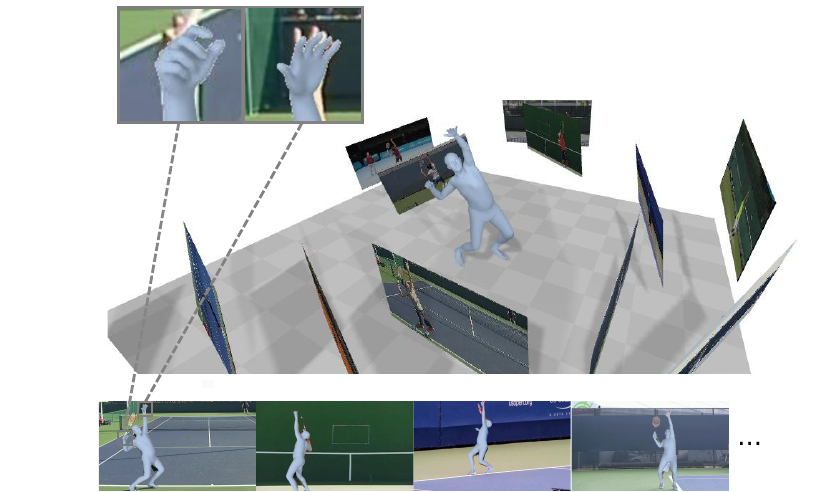}
	\caption{
		This paper proposes a system for motion capture from a set of Internet videos which record different instances of the same action of a person. The videos were recorded at different times and in different scenes (bottom). Our system synchronizes the videos, recover the camera viewpoints, and reconstruct the motion accurately (top). 
	}
	\label{fig:result}
\end{figure}

To make human MoCap a commodity, many monocular motion capture algorithms \cite{zhou2016sparseness,humanMotionKanazawa19,guler2019holopose,xiang2019monocular} have been developed to recover human motion from single RGB videos. Remarkable progress has been made in past years thanks to the advances in deep learning, public datasets on human bodies, and expressive human models \cite{pavlakos2019expressive,loper2015smpl}. However, all these methods take a single video as input. As 3D reconstruction from a monocular image is inherently ill-posed, it is extremely difficult to recover accurate and detailed motion from a single video. Leveraging multiple views can resolve the ambiguity, but calibrated and synchronized multi-view videos are not common. 

Fortunately, we observe that videos of some celebrities doing some specific actions are abundant on the Internet. While those videos were recorded at different times and the motions in these videos are not exactly the same, they encode the same motion characteristics of the person. Compared to a single video, multiple videos provide richer observations about the specific motion. More importantly, the videos are often recorded at different viewpoints which provide multi-view information to help alleviate the 3D ambiguity and self-occlusion issues.

In this paper, we propose to capture human motion from a collection of Internet videos that record different instances of a person's specific performance. 
However, this new problem brings in many challenges that make existing multi-view MoCap algorithms inapplicable: the human motions are not exactly the same among all videos; the videos are unsynchronized; the camera viewpoints are unknown; and the background scenes can be different. 
To solve these challenges, we propose an optimization-based framework that simultaneously solves video synchronization, camera calibration, and human motion reconstruction. More specifically, the proposed system initializes per-frame 3D human pose estimation with a learned 3D pose estimator, synchronizes videos by matching frames based on the 3D pose similarity, and jointly optimizes for camera poses and human motions over all the videos. The motions to be recovered are not assumed exactly the same among videos but modeled by a low-rank subspace. Finally, the motion reconstruction and the pose-based video synchronization are iteratively refined. We also show that the video synchronization can be improved by imposing the cycle consistency constraint among multiple videos. 

In summary, we make the following contributions: 
\begin{itemize}
	\item We introduce the new task of motion capture from a collection of Internet videos that record different instances of a person's certain action, which is unexplored in the literature to our knowledge. 
	\item We develop a new optimization-based framework to solve this new task. Our technical contributions include pose-based video synchronization, low-rank modeling of motions, and joint optimization for synchronization, camera poses and human motion. 
	\item We show that, compared to using single videos, the joint analysis of multiple videos provides richer information to address occlusion and depth ambiguity, even if the videos record different motion instances.
\end{itemize}

\section{Related work}

\paragraph{\bf Single-view Mocap:} There has been remarkable progress on 3D human pose and shape estimation from single images. Many works focus on the skeleton-based 3D human pose estimation, either first estimating 2D pose from images and then lifting it to 3D \cite{moreno20173d,zhou2016sparseness,chen20173d,martinez2017simple,pavllo20193d}, or end-to-end regressing to obtain the 3D pose directly \cite{tome2017lifting,sun2017compositional,tekin2017learning,zhou2017towards,pavlakos2018ordinal,sun2018integral}. In addition, a lot of works propose to estimate the 3D pose and shape involving a parametric model of the human body \cite{anguelov2005scape,loper2015smpl}. Some early works attempt to use the optimization-based methods \cite{sigal2008combined,guan2009estimating,bogo2016keep,lassner2017unite,zanfir2018monocular}, which fit the human model to 2D evidence. More recently, many works attempt to directly regress the model from images with a deep network \cite{kanazawa2018end,omran2018neural,pavlakos2018learning,zanfir2018deep,humanMotionKanazawa19,guler2019holopose,xiang2019monocular}. However, due to the inherent depth ambiguity of single views, the accuracy of these methods is not comparable with the multi-view reconstruction. 

\paragraph{\bf Multi-view Mocap:} Markerless multi-view motion capture has been explored in computer vision for many years. The solutions to this problem are mainly divided into two categories: tracking and pose estimation. Most multi-view tracking methods \cite{gall2010optimization,bo2010twin,lee2010coupled,li20103d,elhayek2015efficient} fit a human body model, e.g., a triangle mesh or a collection of geometric primitives, to image evidence such as keypoints and silhouette. The main difference between them is the type of image evidence and the way to optimize it. However, these tracking based approaches usually require the initialization of the first frame and easily fall into local optima and tracking failures. Hence, more recent works \cite{sigal2012loose,burenius20133d,pavlakos2017harvesting,joo2018total} generally tend to estimate 3D human body based on 2D features detected from images. Burenius et al. \cite{burenius20133d} propose to extend the pictorial structure model to 3D and use it to estimate 3D human skeleton from images. Pavlakos et al. \cite{pavlakos2017harvesting} propose to use a ConvNet for 2D pose estimation and combine with the 3D pictorial structure model to produce 3D pose estimation.  Huang et al. \cite{MuVS:3DV:2017} and Joo et al. \cite{joo2018total} propose to combine statistical body models with a 2D pose estimator and show impressive results. All the above methods assume the multi-view videos are synchronized with known camera parameters. 

There are a few methods \cite{hasler2009markerless,elhayek2012spatio,elhayek2015outdoor,zheng2015sparse,wang2017outdoor,xu2019discrete,saini2019markerless} which attempt to reconstruct the 3D human motion from multiple uncalibrated and unsynchronized videos. Most methods synchronize the videos using additional information, such as audio\cite{hasler2009markerless,elhayek2012spatio}, system time\cite{saini2019markerless}, and flashing a light\cite{wang2017outdoor}, which is unavailable in our scenario. In terms of calibration, many works\cite{hasler2009markerless,elhayek2012spatio,zheng2015sparse,wang2017outdoor,xu2019discrete} assume that the camera parameters are provided or obtain the camera geometry using structure from motion based on the static background, which is inapplicable in our setting where the scenes are totally different. Some works \cite{elhayek2015outdoor,xu2019discrete} also propose to optimize camera parameters and human poses jointly but they assume the motions among videos are exactly the same.

\paragraph{\bf Video alignment:} When the videos are recording the same event, there are many existing methods to address the temporal alignment problem. Early works \cite{caspi2002spatio,tuytelaars2004synchronizing,wolf2006wide,ukrainitz2006aligning} generally assume a linear temporal mapping between videos. More recent works propose non-linear solutions based on handcrafted features \cite{wang2014videosnapping} or learned features \cite{sermanet2018time}. However, for our situation where the videos record similar motions rather than the same event, these approaches are not suitable. Dwibedi et al. \cite{dwibedi2019temporal} propose a self-supervised representation learning method for general video alignment but not tailored for human videos. We will use it as a baseline to evaluate our synchronization component in experiments.

\section{Methods}

Our goal is to reconstruct human motion from multiple videos. Suppose the videos are synchronized, the cameras are calibrated and the motion is the same in these videos, this problem is reduced to a multi-view 3D pose reconstruction problem, which can be solved by first detecting 2D poses in each view and then lifting them to 3D by triangulation. However, this is not the case in our task, where we need to solve video synchronization and motion reconstruction simultaneously with unknown camera geometry, and the motions are similar but not exactly the same across videos.

\begin{figure*}
	\centering
	\includegraphics[width=0.9\linewidth,trim={6cm 6.2cm 6cm 4cm},clip]{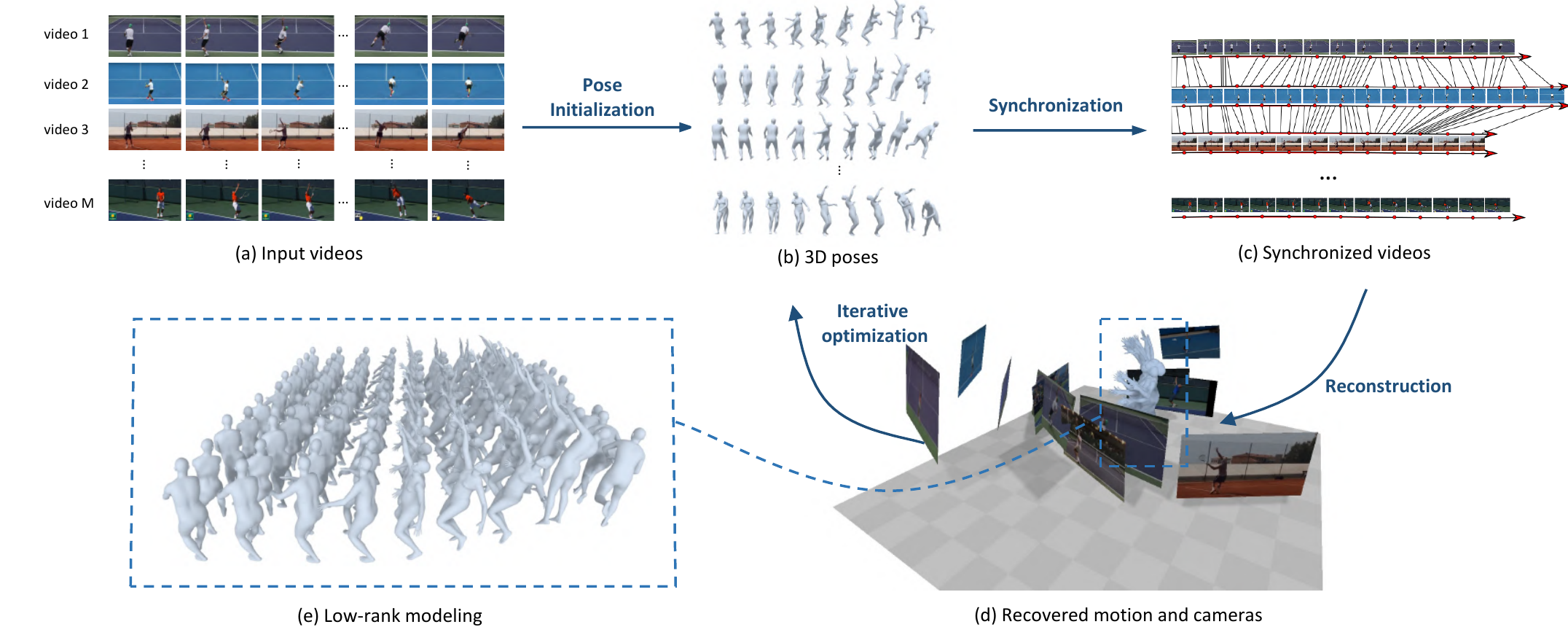}
	\caption{
		\textbf{Overview of our approach.} Given multiple Internet videos of an action (a), an off-the-shelf 3D human pose estimator is used to initialize the 3D pose of each frame (b). Then, the 3D poses are used to synchronize all videos (c), from which the human motion and camera parameters are recovered (d) with the motion variation across videos modeled by a low-rank matrix (e). Finally, the optimized pose estimates are used to refine video synchronization, and video synchronization and motion reconstruction are optimized iteratively. 
	}
	
	\label{fig:pipeline}
\end{figure*}

To solve this challenging problem, we propose an iterative optimization framework that jointly solves synchronization and reconstruction. The intuition is that, if the 3D pose in each video frame is given, we can synchronize videos based on the 3D poses; and if the videos are synchronized, we can recover 3D poses and camera viewpoints from the corresponding frames using multi-view geometry. Figure \ref{fig:pipeline} presents an overview of our approach. We initialize per-frame 3D poses with a CNN-based estimator and iteratively solve synchronization and motion recovery by optimization. In the rest of this section, we first introduce the pose-based video synchronization and then the motion recovery method.   

\subsection{Pose-based video synchronization}\label{sec:align}

In order to leverage multiple views for pose reconstruction, video synchronization is required, i.e., finding the correspondences of frames between videos. However, this is a challenging task because the appearances are very different among videos due to the different background, clothing, and viewpoints. To address this problem, we propose to synchronize videos directly based on 3D human poses seen in the video frames. The initial poses can be obtained by an off-the-shelf pose estimator \cite{humanMotionKanazawa19} and refined after synchronization.

Suppose there are $M$ Internet videos, $N_j$ is the number of frames for video $j$, and $\bm K_{ij} \in \mathbb R^{3 \times J}$ denotes the 3D human pose estimated for the $i$-th frame of video $j$. Then, we can measure the likelihood that two frames correspond to each other (a.k.a affinity) based on the similarity between the estimated 3D human poses. Specifically, we compute the Euclidean distance between each pair of 3D poses aligned by the Procrustes method. Then, we map the reciprocal of distance to a value between $[0,1]$ as the affinity score between two frames. For a pair of videos $j_1$ and $j_2$, we construct an affinity matrix $\bm A_{j_1j_2} \in \mathbb R^{N_{j_1} \times N_{j_2}}$ which consists of all affinity scores between frames of two videos. The correspondences to be estimated can be represented as a partial permutation matrix $\bm X_{j_1j_2} \in \mathbb \{0,1\}^{N_{j_1} \times N_{j_2}}$ and efficiently estimated based on $\bm A_{j_1j_2}$ using an optimal assignment algorithm, e.g., dynamic programming considering the sequential constraint on video frames. 

\begin{figure}[t]
	\centering
	\includegraphics[width=0.6\linewidth,trim={4.6cm 6.5cm 9cm 7cm},clip]{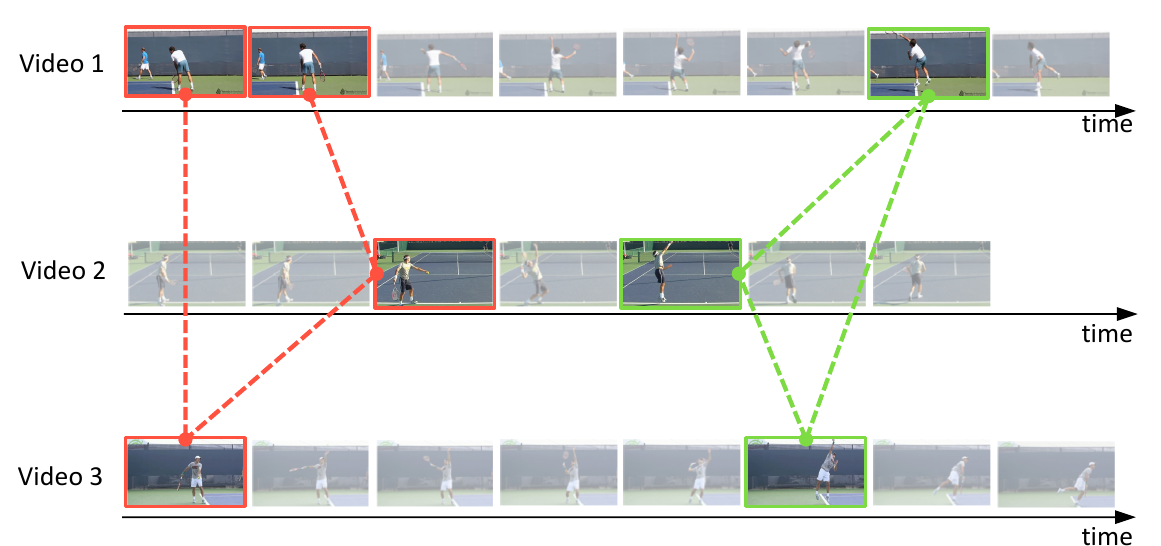}
	\caption{
		\textbf{An illustration of cycle consistency.} The green lines denote a set of consistent correspondences and the red lines show a set of inconsistent correspondences.  
	}
	\label{fig:cycle}
\end{figure}

If we align each pair of videos separately, the resulting correspondences may be inconsistent due to ignoring the cycle consistency constraint. For example, as shown in Figure \ref{fig:cycle}, the correspondences in green are cycle-consistent since they form a closed cycle and the ones in red are inconsistent. Therefore, we can use the cycle consistency constraint to improve the alignment of multiple videos. To achieve this, we adopt the result in prior work \cite{huang2013consistent} that the cycle consistency is equivalent to a low-rank constraint on the correspondence matrix $\bm X$, which is the concatenation of all pairwise permutation matrix:
\begin{equation}\label{eq:big_X}
\bm{X} = 
\left(
\begin{matrix}
\bm{X}_{11} & \bm{X}_{12} & \cdots & \bm{X}_{1M} \\
\bm{X}_{21} & \bm{X}_{22} & \cdots & \bm{X}_{2M} \\
\vdots & \vdots & \ddots & \vdots \\
\bm{X}_{M1} & \bm{X}_{M2} & \cdots & \bm{X}_{MM}
\end{matrix}
\right) \in \mathbb R^{N_a \times N_a}.
\end{equation}
$N_a$ is the number of all frames of all videos.

Therefore, we minimize the following objective function to estimate $\bm X$:
\begin{equation}\label{eq:matching}
\begin{split}
f(\bm X) = \| \bm A - \bm X \|_F^2 + \lambda \cdot rank(\bm X),
\end{split}
\end{equation}
where $\bm A \in \mathbb R^{N_a \times N_a}$ denotes the concatenation of all $\bm A_{j_1j_2}$ similar to the form of $\bm X$, $\lambda$ is the weight of low-rank constraint. This problem can be approximately solved with the convex relaxation algorithms in previous work \cite{dong2019fast,zhou2015multi}. The relaxed solution $\bm X_{j_1j_2}$ is usually not a valid permutation matrix but a real matrix with values in $(0,1)$, which can be regarded as a denoised version of $\bm A_{j_1j_2}$ with cycle consistency.
Finally, to find the frame-to-frame correspondence between video $i$ and video $j$, we use the dynamic time warping algorithm based on the affinity matrix $\bm X_{j_1j_2}$.

\subsection{Motion recovery}\label{sec:recon}

Even if the videos are synchronized, the problem still cannot be treated as a standard multi-view reconstruction problem for the following two reasons. First, the relative camera poses between videos are unknown and cannot be recovered from structure from motion as the scenes in videos are different. Second, the motions in all videos are not exactly the same. To solve the first issue, we directly register cameras with the human body as the reference and recovery the human motion and camera parameters simultaneously. To address the second issue, we propose to model the motion variation among videos by a low-rank subspace. Before we introduce the methods in detail, we first introduce the representation of human motion.  

\paragraph{\bf Motion representation:} For each video, the corresponding 3D human motion is individually represented by a statistical body mesh model SMPL \cite{loper2015smpl} instead of 3D skeleton, since it contains a richer body prior. The SMPL model is parameterized by the pose parameters $\bm \theta \in \mathbb R^{72}$, the shape parameters $\bm{\beta} \in \mathbb R^{10}$, and a root translation $\bm{\gamma} \in \mathbb{R}^3$, and maps a set of parameters to a body mesh denoted by $M(\bm \theta, \bm \beta, \bm \gamma) \in \mathbb R^{3 \times N_v}$ with $N_v=6890$ vertices. A predefined set of 3D body joints $F(\bm \theta, \bm \beta, \bm \gamma) \in \mathbb R^{3 \times J}$ can be generated by linear regression from the mesh vertices, where $J$ denotes the number of 3D joints. The SMPL+H model\cite{MANO:SIGGRAPHASIA:2017} which extends SMPL with hands and SMPL-X model \cite{pavlakos2019expressive} which extends SMPL with face and hands can also be used if the video resolution is sufficient  for OpenPose\cite{openpose} to capture the face and hand motion. Our goal is to recover $\bm \theta_{ij}$, $\bm \beta_{ij}$, and $\bm \gamma_{ij}$, which denote the pose, shape and translation parameters for each frame $i$ of video $j$, respectively. Note that we assume the shape parameters $\bm \beta_{ij}$ remains the same in one video, i.e., $\bm \beta_{ij}= \bm \beta_j$.

\paragraph{\bf SMPL-BA:} We attempt to solve camera parameters and SMPL parameters simultaneously by minimizing the reprojection errors of body keypoints detected in video frames, similar to bundle adjustment in traditional structure from motion. The body keypoints are anchored on the SMPL model. Therefore, we call this procedure SMPL-BA. 

Suppose $\bm R^c_j$ and $\bm T^c_j$ denote the rotation and translation of the camera $j$ in the world coordinate system that defines SMPL, respectively. Then, the reprojection error in SMPL-BA can be written as:
\begin{equation}\label{eq:loss_2d}
L_{2d} = \sum_{i,j,z} c_{ijz}  \rho \left(  \bm W_{ijz} - P\{\bm R^c_jF(\bm \theta_{ij}, \bm \beta_j, \bm \gamma_{ij})_z+\bm T^c_j\} \right) ,
\end{equation} 
where $\bm W_{ijz} \in \mathbb R^{2}$ denotes the $z$-th joint of the estimated 2D pose at $i$-th frame in video $j$ with corresponding confidence $c_{ijz}$ and $P$ denotes the perspective projection. $\rho$ denotes the Geman-McClure robust error function for suppressing noisy detection.

In (\ref{eq:loss_2d}), the camera poses $\bm R^c_j$ and $\bm T^c_j$ are irrelevant to frame index $i$. But in practice the camera may move in each video. To address this issue, we assume that the cameras are only allowed to rotate at fixed camera centers, which is a practical assumption, e.g., in sports broadcasting. Then, we propose to compensate for the camera rotation in each video by warping other frames to the first frame using a homography transformation estimated by feature tracking between frames.

\paragraph{\bf Low-rank modeling of motions:}

When the human motion in each video is not exactly the same, we assume that 3D poses observed in the corresponding frames are very similar which can be approximated by a low-rank matrix:
\begin{equation}\label{eq:low_rank}
rank(\bm \theta_i) \leq s,
\end{equation}
where $\bm \theta_i = [\bm \theta_{i1}^T; \bm \theta_{i2}^T; \cdots ;\bm \theta_{iM}^T] \in \mathbb R^{M \times 72}$ denotes the collection of pose parameters in all videos of frame $i$ and the constant $s$ controls the degree of similarity. Note that each video has its own SMPL parameters. The only constraint that links all videos is the low-rank constraint, which is soft and allows difference among videos.

In addition, we also assume that the 3D trajectories of the root joint of the body should be similar among videos. Suppose the root trajectories in
all videos are denoted by $\bm \gamma=[\bm \gamma_1^T; \bm \gamma_2^T; \cdots; \bm \gamma_M^T] \in \mathbb R^{M \times 3N}$, where $\bm \gamma_j\in \mathbb R^{3N}$ is the trajectory in video $j$ and $N$ is the number of frames. 
Then, the constraint can be written as:
\begin{equation}\label{eq:low_rank_tra}
rank(\bm \gamma) \leq s.
\end{equation}
We set $s$ equal to $1$ or $2$ empirically in our experiments. When the motion variance across videos is large or even there exist outlier videos, a larger $s$ can be used.

\paragraph{\bf Objective function:}

Combining all discussed above, the final objective function to optimize can be written as:

\begin{equation}\label{eq:formulation}
\begin{split}
\min ~ & L_{2d} + \lambda_t L_{temp}, \\
\text{s.t.} ~ \ &rank(\bm \theta_i) \leq s, i=1,2,...,N, \\
&rank(\bm \gamma) \leq s,
\end{split}
\end{equation}
where $L_{temp}$ is a temporal smoothing term with weight $\lambda_t$ to eliminate jittering in motion:
\begin{equation}\label{eq:temporal}
L_{temp} = \sum_{i=1}^{N-1} \| \bm \theta_i - \bm \theta_{i+1} \|^2_F.
\end{equation}

\paragraph{\bf Optimization:}

To simplify the optimization, we introduce two auxiliary variables $\bm Z_i \in \mathbb R^{M \times 72}$ and $\bm Y \in \mathbb R^{M \times 3N}$ to decouple the rank constraints with the objective function:
\begin{equation}\label{eq:rewrite}
\begin{split}
\min ~ &L_{2d} + \lambda_t L_{temp} + \lambda_{r_1} \sum_{i=1}^{N} \|\bm \theta_i-\bm Z_i\|_F^2 + \lambda_{r_2} \|\bm \gamma-\bm Y\|_F^2,\\
\text{s.t.} ~ \ &rank(\bm Z_i) \leq s, i=1,2,\cdots,N, \\
&rank(\bm Y) \leq s
\end{split}
\end{equation}
where $\lambda_{r_1}$ and $\lambda_{r_2}$ are weighting parameters.

The problem in \eqref{eq:rewrite} is highly nonconvex. However, reliable initialization allows us to use local optimization to solve this problem. Specifically, we update each variable alternately while the others remain fixed. The pose $\bm \theta_i$, shape $\bm \beta_j$, and translation $\bm \gamma$ parameters of SMPL can be updated with Gradient Descent. It is a standard low-rank approximation problem to update $\bm Z_i$ and $\bm Y$, which can be solved by SVD analytically. The update of $\bm R^c_j$ and $\bm T^c_j$ can be solved with a perspective-n-point (PnP) algorithm that minimizes reprojection errors over all frames of video $j$. 

\paragraph{\bf Initialization:} We initialize the SMPL parameters for each frame using a pre-trained neural network \cite{humanMotionKanazawa19}, which is further refined by minimizing the reprojection error of 2D keypoints for each frame. Next, the videos are initially synchronized based on the initial pose estimates as introduced in Section \ref{sec:align}. Then, a reference video is selected, whose camera coordinate system is regarded as the world frame. Note that the initial SMPL model in each video is defined in the coordinate system of the respective camera. Therefore, the relative camera poses between two videos can be initialized by rigidly aligning the SMPL models, assuming the SMPL pose parameters are the same between videos. When intrinsics are unknown, we set the focal length to be a large constant, approximating a weak-perspective camera model. In this way, the camera poses can be initialized.

\subsection{Iterative optimization}\label{sec:iter}

The video synchronization in the first iteration may not be very accurate based on initial pose estimates. Therefore, we propose to refine the synchronization based on the optimized poses. More specifically, the affinity matrix $\bm A$ in Section \ref{sec:align} is updated with the optimized poses given by the SMPL-BA and the frame correspondences are re-computed using the new affinity matrix. Then, the SMPL-BA is computed again with the updated synchronization. Both synchronization and reconstruction benefit from each other in iterative optimization, which will be experimentally demonstrated in Section \ref{sec:ablation}.

\section{Experiments}

\subsection{Motion Capture from Internet videos} \label{sec:dataset}

\begin{figure}[t]
	\centering
	\begin{minipage}{.47\textwidth}
		\centering
		\includegraphics[width=1.45\linewidth,trim={5.5cm 3cm -1.5cm 1cm},clip]{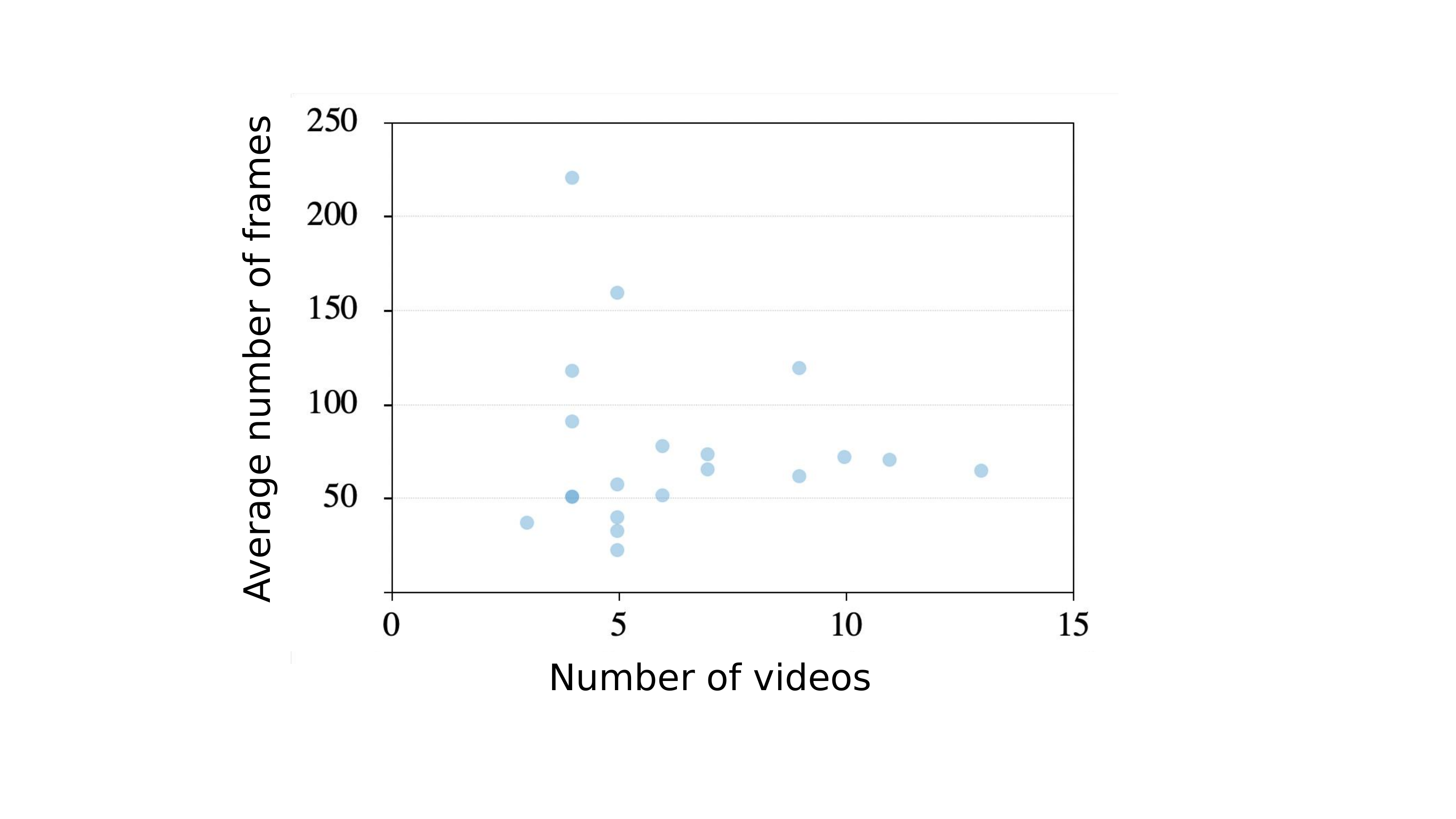}
		\captionof{figure}{\textbf{Collected Internet video dataset}. Each point denotes one action.}
		\label{fig:infor}
	\end{minipage} \quad%
	\begin{minipage}{.48\textwidth}
		\centering
		\includegraphics[width=1\linewidth,trim={2cm 3cm 4cm 4cm},clip]{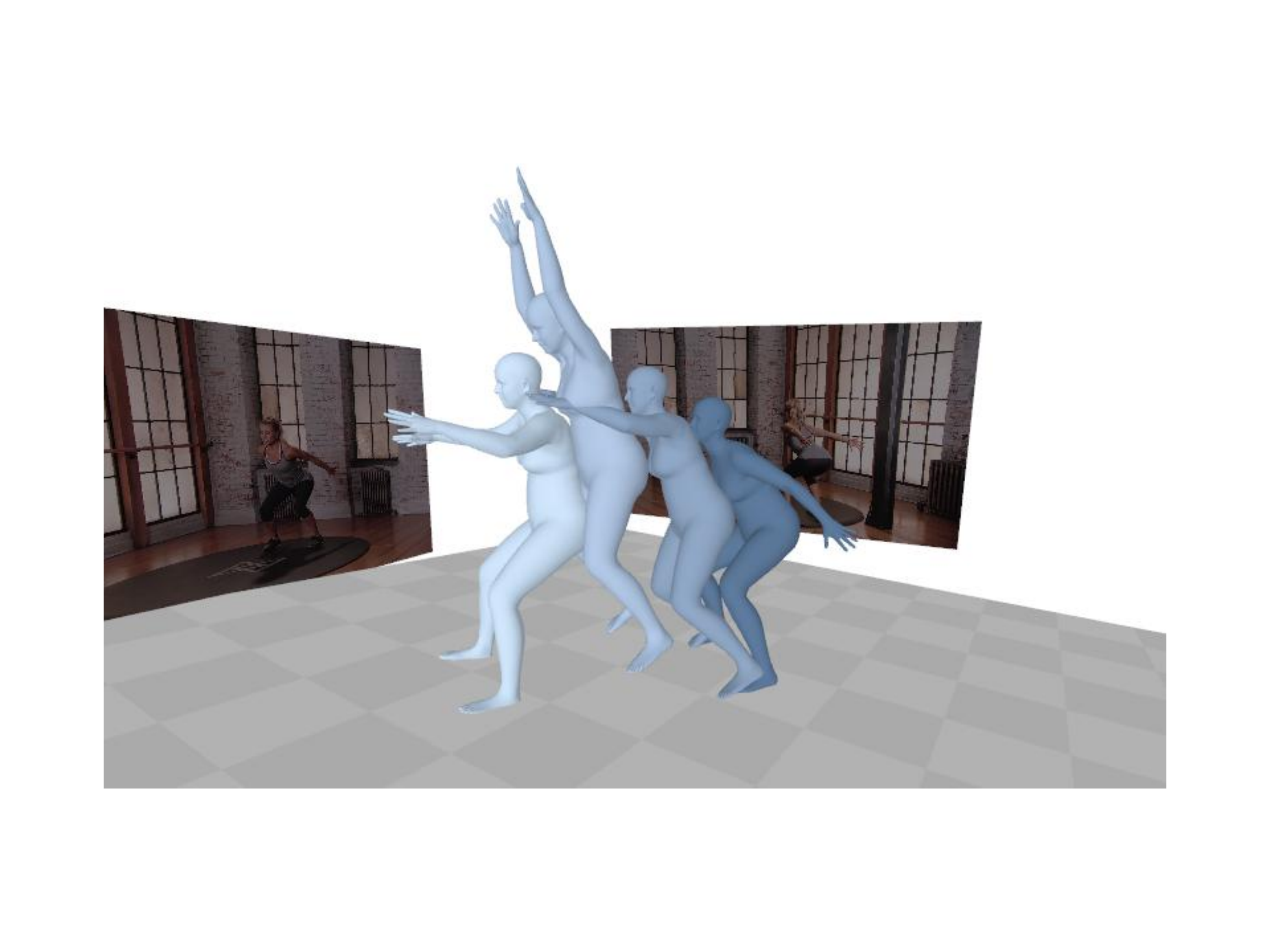}
		\captionof{figure}{
			\textbf{Trajectory recovery}. Our approach is able to recover the absolute 3D trajectory of human motion. The brightness of human mesh indicates the chronological order. 
		}
		\label{fig:traj}
	\end{minipage}
\end{figure}

There is no existing dataset for our task. Therefore, we collect a new dataset that consists of 20 actions of various actors, such as tennis serves, yoga and Tai Chi. Take tennis serves as an example. We download the publicly available videos of some tennis players from YouTube, and manually crop the videos roughly to obtain a set of video clips of serves for each player. Figure \ref{fig:infor} shows the statistics of the number of videos and average number of frames for each action. The dataset is available at \url{https://github.com/zju3dv/iMoCap}.

\begin{figure*}
	\centering

	\centering
	\includegraphics[width=1\linewidth,trim={1cm 0.5cm 2cm 0.5cm},clip]{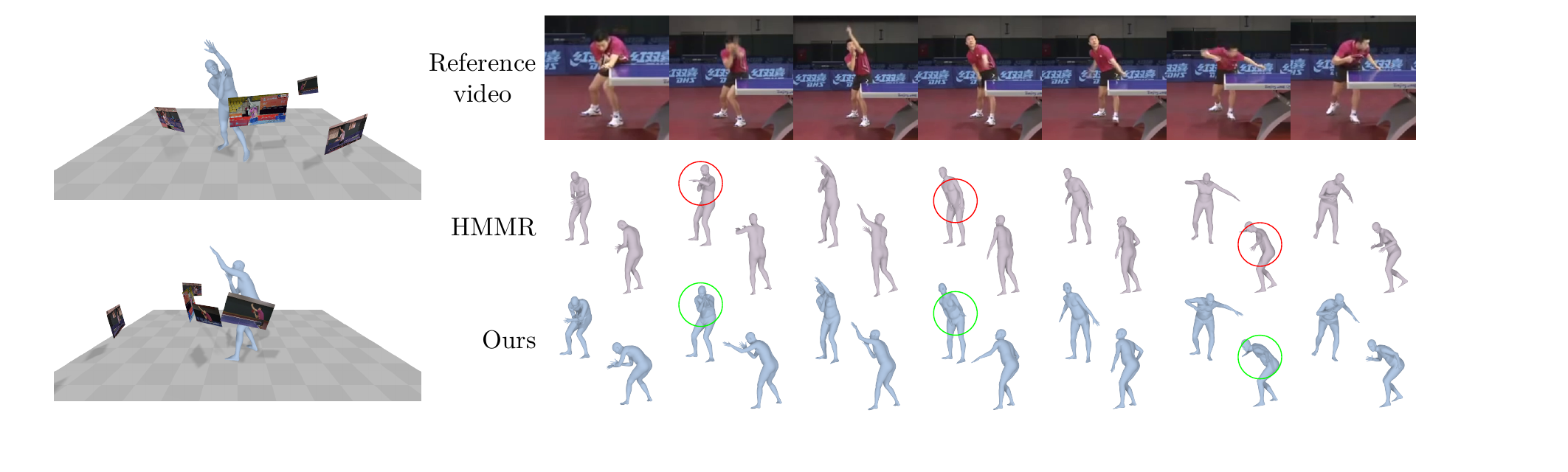}
	\includegraphics[width=1\linewidth,trim={1cm 0.4cm 2cm 0.3cm},clip]{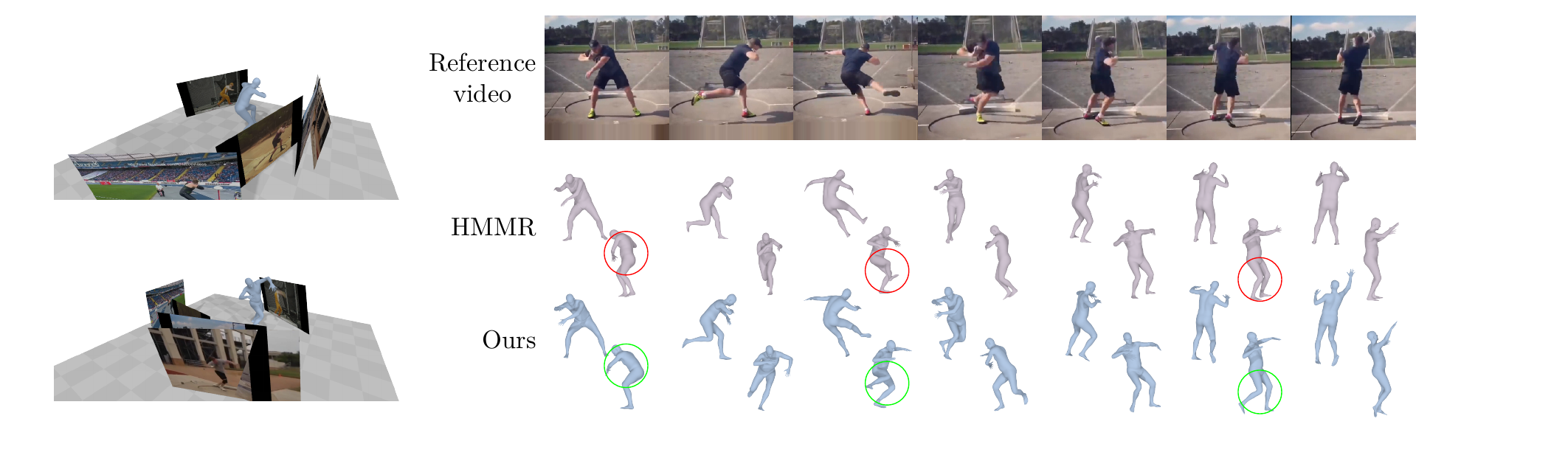}
	\includegraphics[width=1\linewidth,trim={1cm 0.4cm 2cm 0.3cm},clip]{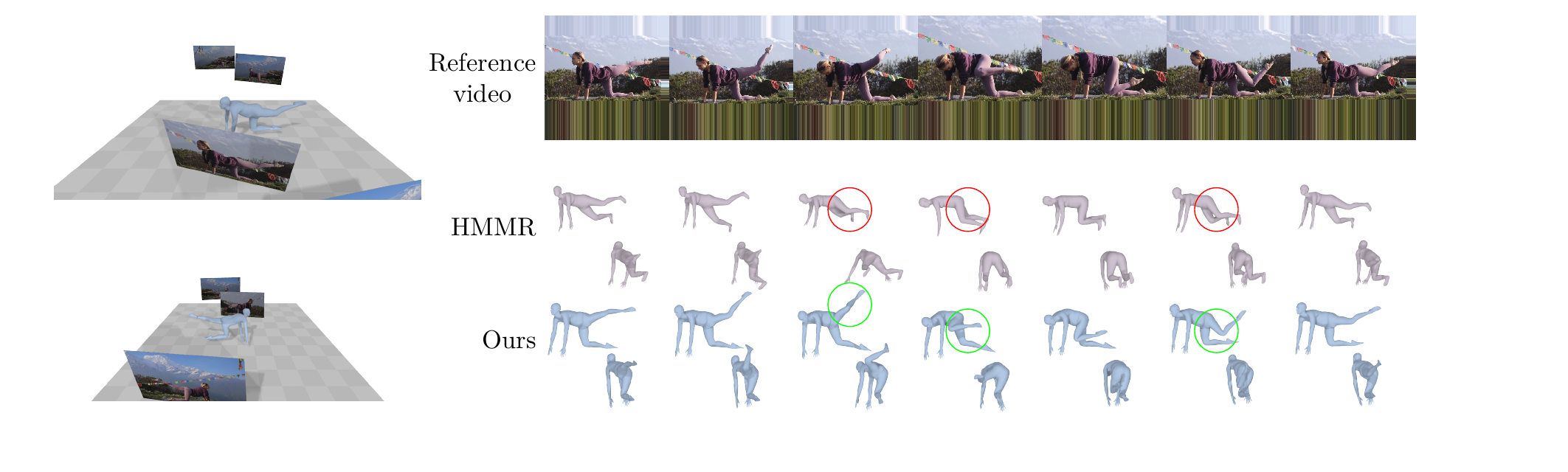}
	\includegraphics[width=1\linewidth,trim={1cm 0cm 2cm 0.3cm},clip]{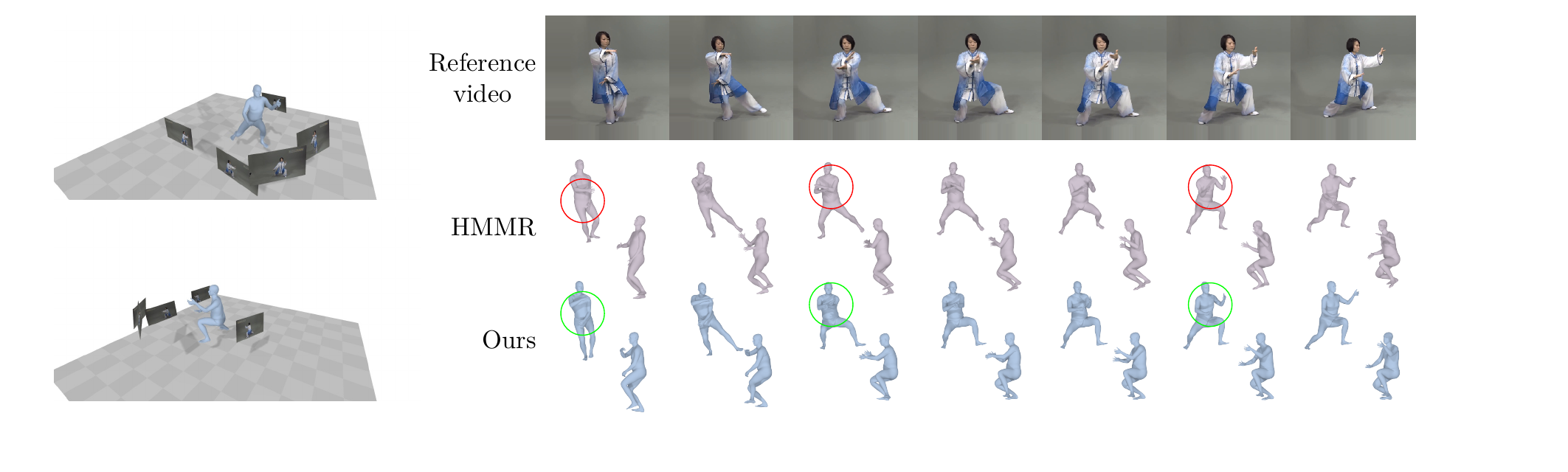}
	\includegraphics[width=0.55\linewidth,trim={0.5cm 6cm 7cm 8cm},clip]{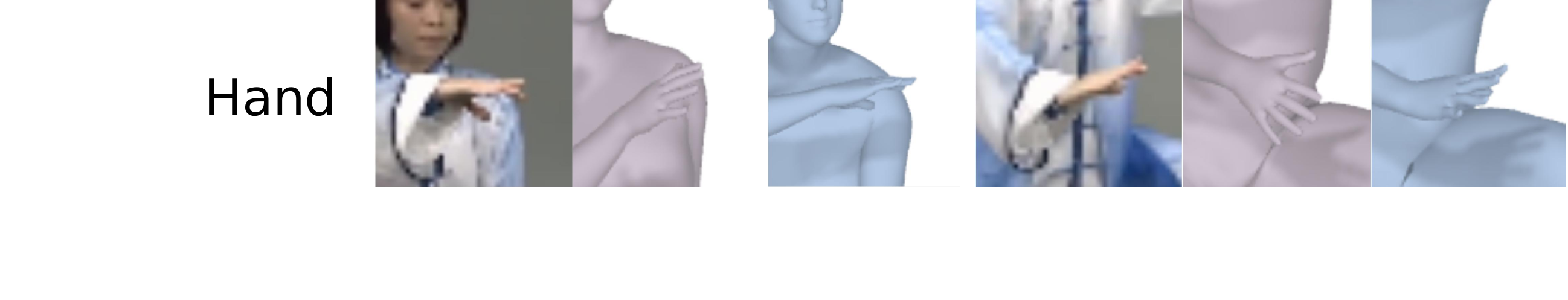}

	\caption{
		\textbf{Results on Internet videos} of table tennis serves, shotput, yoga, and Tai Chi(with hands motion). The left images present the reconstructed human motion and camera positions visualized in two viewpoints. On the right, we present some frames of the reference video and corresponding motion capture results from HMMR \cite{humanMotionKanazawa19} and our method. Red and green cycles emphasize some representative differences between the results from HMMR and our method.
	}
	\label{fig:qual}
\end{figure*}

We apply the proposed approach on each action of this dataset to recover the corresponding human motion. Some representative results are visualized in Figure \ref{fig:qual}, which shows that the proposed approach is able to recover 3D human motion as well as camera geometry from these videos, even if they were recorded at different times. 
These videos record the action from very different viewpoints and therefore provide multi-view constraints to help alleviate the depth ambiguity and self-occlusion issues that often occur for single-view estimation. Consequently, compared to the monocular motion capture algorithm \cite{humanMotionKanazawa19}, our approach produces much more detailed and faithful motion, as indicated by the circles in Figure \ref{fig:qual}. In addition, with the multi-view constraint, our method is also able to recover an accurate 3D trajectory of the body as shown in Figure \ref{fig:traj}, which is infeasible for monocular motion capture algorithms. Note that, the proposed approach can be easily extended to  hand motion recovery if the 2D hand pose estimation is available as shown in Figure \ref{fig:result} and \ref{fig:qual} (Tai Chi). We find that most of the failure cases are because of failed 2D pose estimation. Also, when the viewpoints of videos are similar, the depth ambiguity cannot be resolved even if multiple videos are used. \emph{More qualitative results and video demonstrations are available in the supplementary material.}

\begin{figure*}
	\centering
	\includegraphics[width=0.88\linewidth,trim={0.5cm 5cm 0.9cm 1.7cm},clip]{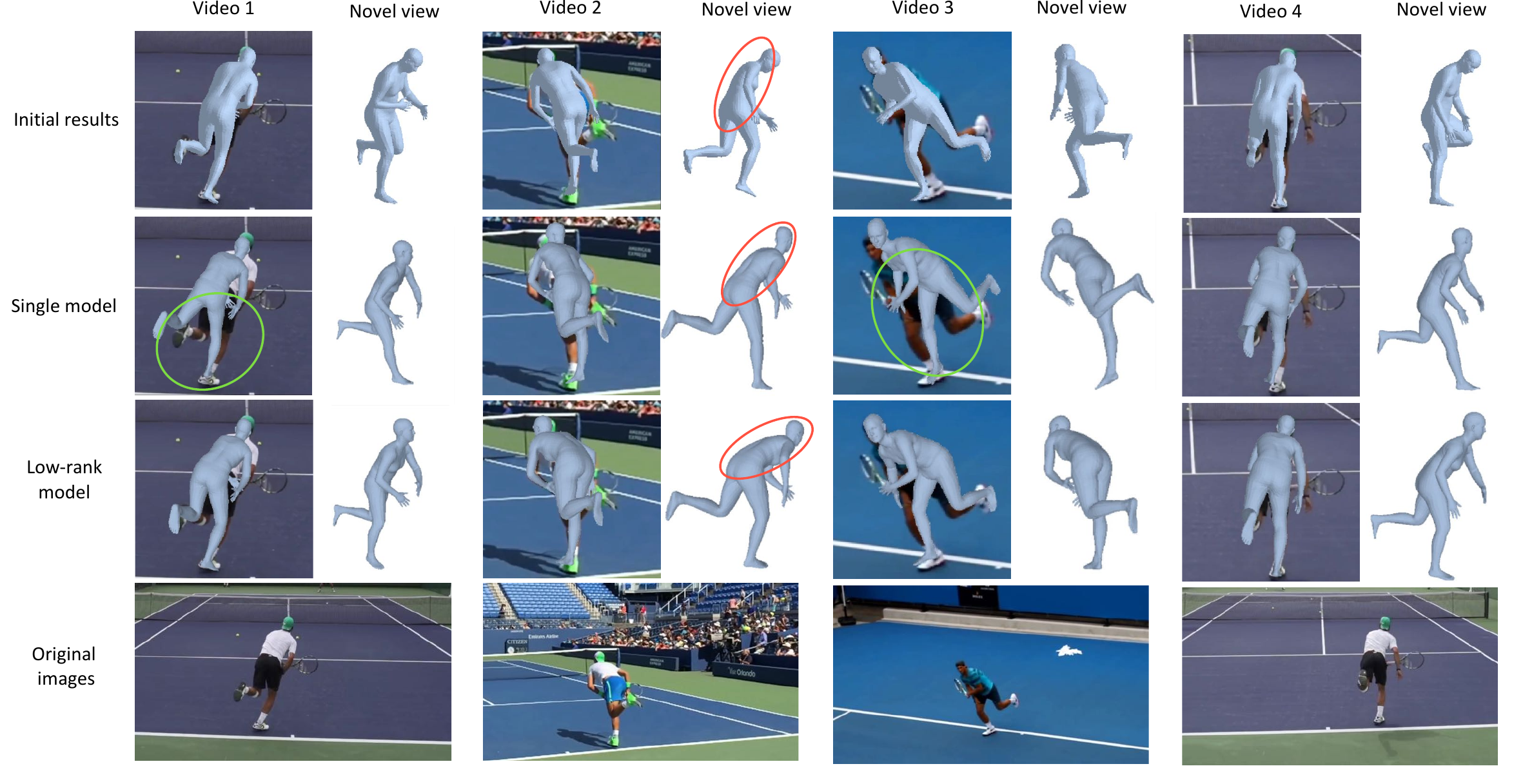}
	\caption{
		\textbf{Effect of low-rank modeling}. The projected mesh of a single model is less consistent with 2D evidence (red marks). Low-rank modeling is able to capture differences among videos and recover more accurate motion such as the curved back of the performer (green marks).
	}
	
	\label{fig:ablation}
\end{figure*}

Since the motion in all the videos is not exactly the same, 
each video has its own SMPL parameters with a low-rank constraint to make the parameters correlated among multiple videos. 
An alternative is to assume the motions are all the same and use a single model with the same set of SMPL parameters for all videos. We provide a qualitative comparison in Figure \ref{fig:ablation}. More specifically, we reproject the initial 3D mesh, the 3D mesh reconstructed by our low-rank model, and the 3D mesh reconstructed by the single model to images and compare them in terms of 2D consistency. The results show that the projected mesh using the single model is less consistent with 2D evidence as shown in the red circles, which suggests that the single model cannot model the motion difference among videos. Low-rank modeling is able to capture more detailed motion, such as the curved back of the performer, as shown in the green circles. Overall, low-rank modeling recovers more detailed and natural motion than the results of a single model and initial monocular results.

\subsection{Quantitative evaluation} \label{sec:ablation}

\begin{figure}[t]
	\centering
	\includegraphics[width=1\linewidth,trim={3cm 7.9cm 1cm 6cm},clip]{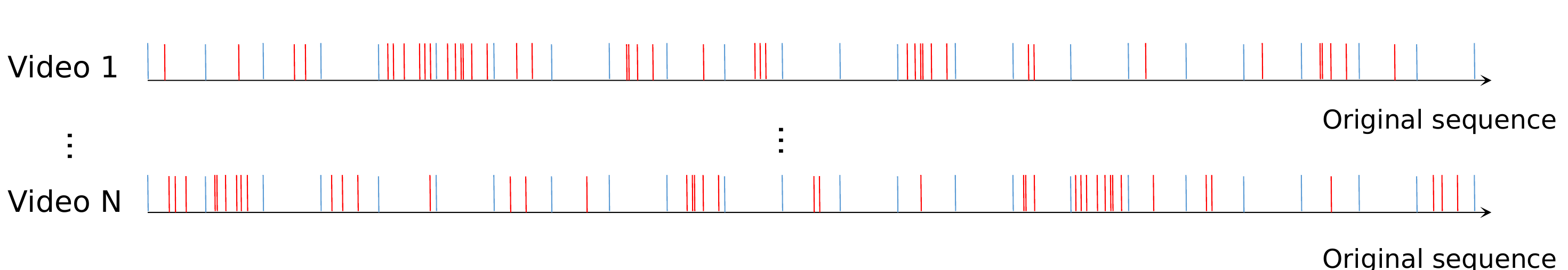}
	\caption{
		\textbf{Dataset generation for quantitative analysis}.
		We edit the videos in Human3.6M to simulate the unsynchronized scenario. As the dataset is large, we only select a few actions, i.e., SittingDown, Sitting, Smoking, Photo and Phoning. For each action, we first sample $N_{s1}$ frames at equal intervals (\textbf{blue lines}) from each video, which results in $N_{s1}-1$ segments. Then we randomly choose $N_{s2}$ segments and randomly sample $N_{s3}$ frames (\textbf{red lines}) from each selected segment. In our experiments, we set $N_{s1}=150, N_{s2}=50$ and for each segment $N_{s3}$ is a variable value randomly selected from 1 to the length of the segment. The dataset is available at \url{https://github.com/zju3dv/iMoCap}.
	}
	\label{fig:h36m}
\end{figure}

While we have collected a dataset of Internet videos to demonstrate the qualitative performance of our system, \textit{quantitative} evaluation is difficult due to the lack of 3D ground truth, a similar case for most prior work on reconstruction from Internet data. For quantitative analysis, we synthesize a dataset using existing datasets \cite{ionescu2013human3} with ground-truth annotations. We select some challenging sequences in the Human3.6M dataset \cite{ionescu2013human3} and modify the data to simulate the unsynchronized and uncalibrated scenario. Please refer to Figure \ref{fig:h36m} for details. We would like to note that the purpose of evaluation on Human3.6M is not to compare against existing methods in the standard Human3.6M setting, but to provide an ablative analysis of our system when solving the proposed problem.

\paragraph{\bf Video synchronization:} As described in section \ref{sec:align}, we propose a pose-based video synchronization method to address the different appearances among videos, i.e., background, clothing, and viewpoints. We also impose the cycle consistency constraint to improve the synchronization. Here, we compete with some baselines and we use the standard video alignment metric to measure the alignment of two videos. In particular, for each frame of non-reference video $v_i$, we compute the frame distance between the matched frame and the ground truth position in reference video $v_0$ and normalize it by the video clip length.

We first propose a simple alternative to use the DP algorithm to quantize the original affinity matrix directly. The result of this baseline method (`No cycle-consis') is shown in Table \ref{tb:ablative_sync}. The results show that imposing the cycle consistency constraint can reduce the alignment error of video synchronization significantly.

Another baseline is a recent self-supervised representation learning method \cite{dwibedi2019temporal} based on the cycle consistency loss to align  videos. Their network is retrained on the evaluated sequences and the alignments are obtained by the dynamic time warping algorithm on the features. The results of their method (`TCC') on the dataset are presented in Table \ref{tb:ablative_sync}. The results show that our 3D pose based method outperforms the generic method by a large margin in our case. 
\begin{table}[t]
	\begin{center}
	\caption{\textbf{Quantitative analysis of synchronization.} `No cycle-consis' denotes our synchronization method without cycle consistency constraint. `TCC' represents the general video synchronization method \cite{dwibedi2019temporal} based on representation learning.  }\label{tb:ablative_sync}
		\setlength{\tabcolsep}{4pt}
		\begin{tabular}{rc}\cline{1-2}
			Method & Synchronization error  \\\hline
			Ours & {\bf0.77\%}    \\
			No cycle-consis & {1.19\%}  \\\hline
			TCC \cite{dwibedi2019temporal} & 11.24\%  \\\hline

		\end{tabular}
	\end{center}

\end{table}

\paragraph{\bf Reconstruction:} We evaluate the motion reconstruction quantitatively. To evaluate 3D joint error, we use the standard metric, i.e., the mean per joint position error (MPJPE) and the error after rigid alignment with the ground truth (P-MPJPE).

As videos are unsynchronized and uncalibrated, none of the existing multi-view MoCap methods is applicable to the proposed problem. Monocular MoCap methods are the only applicable alternatives. We compare with the state-of-the-art monocular method HMMR \cite{humanMotionKanazawa19} and the results are shown in Table \ref{tb:ablative_hmmr}. Our method significantly reduces the reconstruction error compared to HMMR, which shows the benefit of using multiple videos.

\begin{table}[t]
	\begin{center}
		\caption{\textbf{Quantitative analysis of reconstruction.} `HMMR' denotes the state-of-the-art monocular motion capture method \cite{humanMotionKanazawa19}. }\label{tb:ablative_hmmr}
		\setlength{\tabcolsep}{4pt}
		\resizebox{0.7\linewidth}{!}{
		\begin{tabular}{rcc}\cline{1-3}
			Method  & MPJPE (mm) & P-MPJPE (mm)  \\\hline
			HMMR  & 109.80 & 78.26 \\
			Ours+generic 2D, 4 videos   & {76.48} & 53.34 \\\hline
			Ours+fine-tuned 2D, 1 video~ & 80.65 & 62.58 \\
			Ours+fine-tuned 2D, 2 videos  & 78.45 & 59.42 \\
			Ours+fine-tuned 2D, 3 videos  & 71.48 & 53.77 \\
			Ours+fine-tuned 2D, 4 videos  & {\bf 66.53}  & {\bf 50.33} \\\hline

		\end{tabular}
		}
	\end{center}

\end{table}

Note that we use a generic 2D pose detector \cite{fang2017rmpe} which is not fine-tuned on Human3.6M. With no fine-tuning, we wish to evaluate the generalization ability of our system when applied to unseen and challenging videos. We also report the results with a fine-tuned 2D pose detector\cite{Chen2018CPN} in Table \ref{tb:ablative_hmmr}, which show that using a fine-tuned detector significantly reduces the reconstruction error. 

In addition, we validate the influence of the number of videos. We report the reconstruction accuracy with various numbers of videos in Table \ref{tb:ablative_hmmr}. The results show that more videos improve the accuracy of reconstructed motion. As for Internet videos, while more views generally improve the results, we empirically find that three or four videos are sufficient in most cases.

\paragraph{\bf \bf Iterative optimization:} Our approach iteratively optimizes video synchronization and reconstruction to let them benefit from each other. `No iter-opt' in Table \ref{tb:ablative_iter} indicates the result without such an iterative optimization, which shows that iterative optimization reduces both alignment and reconstruction errors.

\begin{table}[t]
	\begin{center}
		\caption{\textbf{Quantitative analysis of iterative optimization.} `No iter-opt' denotes our method without iterative optimization of synchronization and reconstruction.}\label{tb:ablative_iter}
		\setlength{\tabcolsep}{4pt}
		\begin{tabular}{rccc}\cline{1-4}
			Method & Synchronization error & MPJPE (mm) & P-MPJPE (mm)  \\\hline
			Ours & {\bf0.77\%} & {\bf 66.53}  & {\bf 50.33} \\
			No iter-opt  & {0.91\%} & {71.49} & 54.12 \\\hline

		\end{tabular}
	\end{center}

\end{table}

\section{Summary}

In this paper, we demonstrated the potential of leveraging multiple Internet videos to recover accurate and detailed human motion, which in a long-term perspective opens up the possibility of collecting high-quality and diverse human motion data for free from existing Internet videos. 
Unlike standard multi-view motion capture, in this new task the human motions are not exactly the same among all videos; the videos are unsynchronized; the camera viewpoints are unknown; and the background scenes can be different. All these challenges make existing multi-view motion capture algorithms inapplicable. To address all challenges above, we proposed (1) low-rank modeling of motions to handle motion variation among videos; (2) pose-based multi-video synchronization and calibration; and (3) \emph{most importantly} a unified optimization-based framework to solve the entire problem, which doesn't treat synchronization, calibration and motion recovery as separate tasks, but integrates them in a single optimization problem. Both qualitative and quantitative results demonstrated the effectiveness of the proposed approach. Please see the supplementary material for more video demonstrations.

\paragraph{\bf \bf Acknowledgement:} The authors would like to acknowledge support from NSFC (No. 61806176) and Fundamental Research Funds for the Central Universities (2019QNA5022).

\clearpage
\bibliographystyle{splncs04}
\bibliography{egbib}
\end{document}